\documentclass[11pt]{article}

\usepackage[preprint]{acl}

\usepackage{times}
\usepackage{latexsym}

\usepackage[T1]{fontenc}

\usepackage[utf8]{inputenc}

\usepackage{microtype}

\usepackage{inconsolata}

\usepackage{graphicx}

\usepackage{microtype}
\usepackage{hyperref}
\usepackage{url}
\usepackage[T1]{fontenc}
\usepackage{booktabs}
\usepackage{listings}
\usepackage{xcolor} 
\usepackage{CJKutf8}
\usepackage{lineno}

\usepackage{amsmath,amsfonts,bm}

\def\eqref#1{equation~\ref{#1}}

\def\1{\bm{1}}

\def\vf{{\bm{f}}}
\def\vg{{\bm{g}}}

\def\vs{{\bm{s}}}

\def\vx{{\bm{x}}}
\def\vy{{\bm{y}}}

\DeclareMathAlphabet{\mathsfit}{\encodingdefault}{\sfdefault}{m}{sl}
\SetMathAlphabet{\mathsfit}{bold}{\encodingdefault}{\sfdefault}{bx}{n}

\usepackage{subcaption}
\usepackage{wrapfig}
\usepackage{enumitem}
\usepackage{graphicx}
\usepackage{xspace}
\usepackage{CJKutf8}
\newcommand{\method}{RASST\xspace}
\usepackage[most]{tcolorbox}
\usepackage{multirow}

\usepackage{CJKutf8}
\usepackage{tabularx}

\definecolor{darkblue}{rgb}{0, 0, 0.5}
\hypersetup{colorlinks=true, citecolor=darkblue, linkcolor=darkblue, urlcolor=darkblue}

\usepackage{xcolor}

\title{\method: Retrieval-Augmented Simultaneous Speech Translation}

\author{
 \textbf{Jiaxuan Luo\textsuperscript{1}},
 \textbf{Siqi Ouyang\textsuperscript{2}},
 \textbf{Jiaxing Xu\textsuperscript{2}},
 \textbf{Lei Li\textsuperscript{2}}
\\
 \textsuperscript{1}Johns Hopkins University,
 \textsuperscript{2}Carnegie Mellon University
\\
 \texttt{jluo50@jh.edu},
 \texttt{siqiouya@andrew.cmu.edu}
}

\begin{document}
\maketitle

\begin{abstract}

Simultaneous speech translation produces target text incrementally from partial speech input.
Recent speech large language models have markedly improved SST quality but still struggle with rare and domain-specific terminology. Retrieval augmentation has helped in automatic speech recognition and neural machine translation, but extending it to SST is non-trivial: retrieval must be fast and accurate under partial speech, and the model must decide whether and when to apply retrieved terms during incremental generation.
We propose \textbf{R}etrieval-\textbf{A}ugmented \textbf{S}imultaneous \textbf{S}peech \textbf{T}ranslation (\method), which addresses both challenges. 
For accurate cross-modal retrieval under partial input, \method trains a lightweight speech–text retriever that produces chunkwise terminology hints for the Speech LLM via multi-scale retrieval. To use these hints correctly, we synthesize training data that teaches the Speech LLM to decide whether and when to apply each retrieved term. Experiments on ACL 60/60 dev set and the ESO test set show that \method improves terminology accuracy by nearly 40\% and overall translation quality by up to 3 BLEU points, with negligible computational overhead.

\end{abstract}
\section{Introduction}

Simultaneous Speech Translation (SST) aims to generate translations incrementally as partial speech input is received. 
It has wide applications in international conferences and cross-lingual conversations~\citep{ma-etal-2020-simulmt, ren-etal-2020-simulspeech}. 
Recently, Large Language Models (LLMs) have shown strong potential as backbones for SST systems~\citep{wang-etal-2025-conversational, ouyang-etal-2025-infinisst, fu2025efficientadaptivesimultaneousspeech, cheng2025seedliveinterpret20endtoend}. 
However, they still struggle to accurately translate domain-specific terminologies, including technical jargon, person and location names, and abbreviations.
\begin{CJK*}{UTF8}{gkai}
As shown in Table~\ref{tab:intro}, a plain SST system translates "Shinzo Abe" incorrectly as "新三·阿贝", which is a phonetic transliteration rather than the official translation of the politician's name. 
\end{CJK*} 

Human simultaneous interpreters often rely on external glossaries to translate specialized terminology in real time~\citep{jbp:/content/journals/10.1075/intp.15.1.04jia}, and retrieval augmentation has been explored widely in automatic speech recognition (ASR) and neural machine translation (NMT) to improve terminology handling as the glossary size grows~\citep{khandelwal2021nearest,9687898,jayanthi-etal-2023-retrieve,li-etal-2024-optimizing-rare,li-etal-2025-leveraging,wu-etal-2025-locate,gong25_interspeech,kong25_interspeech}. 
However, retrieval-augmented SST poses unique challenges. Beyond requiring a cross-modal retriever that is fast and accurate under partial speech input, SST translations, unlike ASR transcripts which are time-aligned with the speech, lag behind the speech. The model must therefore decide not only whether to use a retrieved term, but also when to apply it during incremental translation.

\begin{CJK*}{UTF8}{gkai}
\begin{table}[t]
    \centering
    \small
    \begin{tabularx}{\linewidth}{X}
    \toprule
        \textbf{Source:} \textcolor{blue!70!black}{Shinzo Abe} served as the Prime Minister of Japan. \\ \midrule
        \textbf{Reference:} \textcolor{green!70!black}{安倍晋三}曾担任日本首相。\\\midrule
        \textbf{Plain SST:} \textcolor{red}{新三·阿贝}曾担任日本首相。 \\ \midrule
        \textbf{\method:} \textcolor{green!70!black}{安倍晋三}曾担任日本首相。\\ \bottomrule
    \end{tabularx}
    \caption{A case showing plain SST without retrieval fails to translate terminology correctly while SST with retrieval is able to do so. English (\textcolor{blue!70!black}{blue}), correct Chinese terms (\textcolor{green!70!black}{green}), and incorrect ones (\textcolor{red}{red}). }
    \label{tab:intro}
\end{table}
\end{CJK*}


To address these challenges, we propose \textbf{R}etrieval-\textbf{A}ugmented \textbf{S}imultaneous \textbf{S}peech \textbf{T}ranslation (\method). \method trains a lightweight speech--text retriever on variable-duration speech contexts with a contrastive loss, and performs multi-scale inference for accurate retrieval under partial input. We further synthesize training data that teaches the Speech LLM to select relevant retrieved terms and decide when to introduce them during generation, conditioned on the prior speech context and previously generated translation. Experiments on the ACL 60/60 dev set and ESO test set show that \method improves terminology translation accuracy by nearly 40\% percent and BLEU by up to 3 points on three language directions (En$\rightarrow$Zh/De/Ja), while adding negligible computational overhead.


\section{Related Works}

\paragraph{SST with LLM} Recent work shows that simultaneous speech translation (SST) achieves markedly higher quality when large language models (LLMs) serve as the backbone~\citep{ahmad-etal-2024-findings}. \citet{koshkin-etal-2024-transllama} adapt an LLM to SST via finetuning on synthetic translation trajectories, and \citet{ouyang-etal-2025-infinisst} scale this idea with an interleaving architecture for unbounded speech streams. \citet{fu2025efficientadaptivesimultaneousspeech} explore alternative training strategies and data-synthesis pipelines, while \citet{guo2025streamuniachievingstreamingspeech} unify streaming transcription and translation with a truncation mechanism that prunes history based on transcribed inputs. However, these systems still struggle with domain-specific terminology and \method addresses this by integrating a terminology retriever directly into the SST process.

\paragraph{Retrieval-Augmented NMT} Retrieval is an effective remedy for translating rare words, named entities, and domain-specific terminology in NMT. Early approaches retrieve translation pieces from similar training pairs~\citep{zhang-etal-2018-guiding}, mine bilingual terminology with bilingual embeddings~\citep{huck-etal-2019-better}, or fuse dictionary entries into NMT via a Pointer-Disambiguator-Copier~\citep{zhang-etal-2021-point}. kNN-MT~\citep{khandelwal2021nearest} augments NMT with nearest-neighbor lookup over a large datastore, and follow-up work improves its robustness~\citep{jiang-etal-2021-learning} and efficiency~\citep{meng-etal-2022-fast,zhu-etal-2023-ink}. Other approaches extract and attend over entity candidates~\citep{zeng-etal-2023-extract} or combine retrieval and demonstration with LLMs~\citep{li-etal-2025-leveraging}. For speech translation, \citet{li-etal-2024-optimizing-rare} prepend retrieved speech segments as demonstrations, and \citet{wu-etal-2025-locate} use sliding-window retrieval to inject terms into a speech-LLM prompt. All of these target offline translation, whereas \method integrates retrieval directly into incremental streaming translation.

\paragraph{Contextual-Biasing ASR} Recognizing rare words and domain-specific terminology, commonly known as contextual biasing, is also well studied in ASR. Early work splits into two lines: joint optimization of the ASR model with contextual n-gram embeddings~\citep{CLAS,jain20_interspeech}, and shallow fusion of biased words into end-to-end models~\citep{shallow_fusion}, later extended with prefix-trie-based deep biasing~\citep{le21_interspeech} and unified via deep shallow fusion and trie-constrained biasing~\citep{9383560}. \citet{9687898} support runtime addition of context words through an external memory accessed at decoding. To scale biasing lists from hundreds to tens of thousands of entries, subsequent work leverages neural associative memories~\citep{9747726,10023323,wu23e_interspeech,wu-etal-2024-deferred} and retrieval augmentation~\citep{9413756,10095857,jayanthi-etal-2023-retrieve}. 
More recent efforts build on speech foundation models. \citet{10832154} use coarse CTC decoding to filter hotwords into an LLM-based ASR prompt, and \citet{gong24b_interspeech} inject bias words into a SpeechLLM via prompt formats and a biasing-fusion network. BR-ASR~\citep{gong25_interspeech} decouples bias retrieval from ASR decoding through contrastive learning and curriculum training, scaling to 200k entries at 20ms retrieval latency, while GLCLAP~\citep{kong25_interspeech} retrieves bias words directly from speech via global-local contrastive language-audio pre-training, enabling plug-in biasing on frozen models like Whisper. Finally, PAC~\citep{11462669} combines pronunciation-guided context learning with pronunciation-discriminative reinforcement learning. 
Unlike ASR, where the transcript is time-aligned with the speech, SST translations lag behind the input, so the model must therefore decide whether and when to apply each retrieved term.
\begin{figure*}
    \centering
    \includegraphics[width=0.9\linewidth]{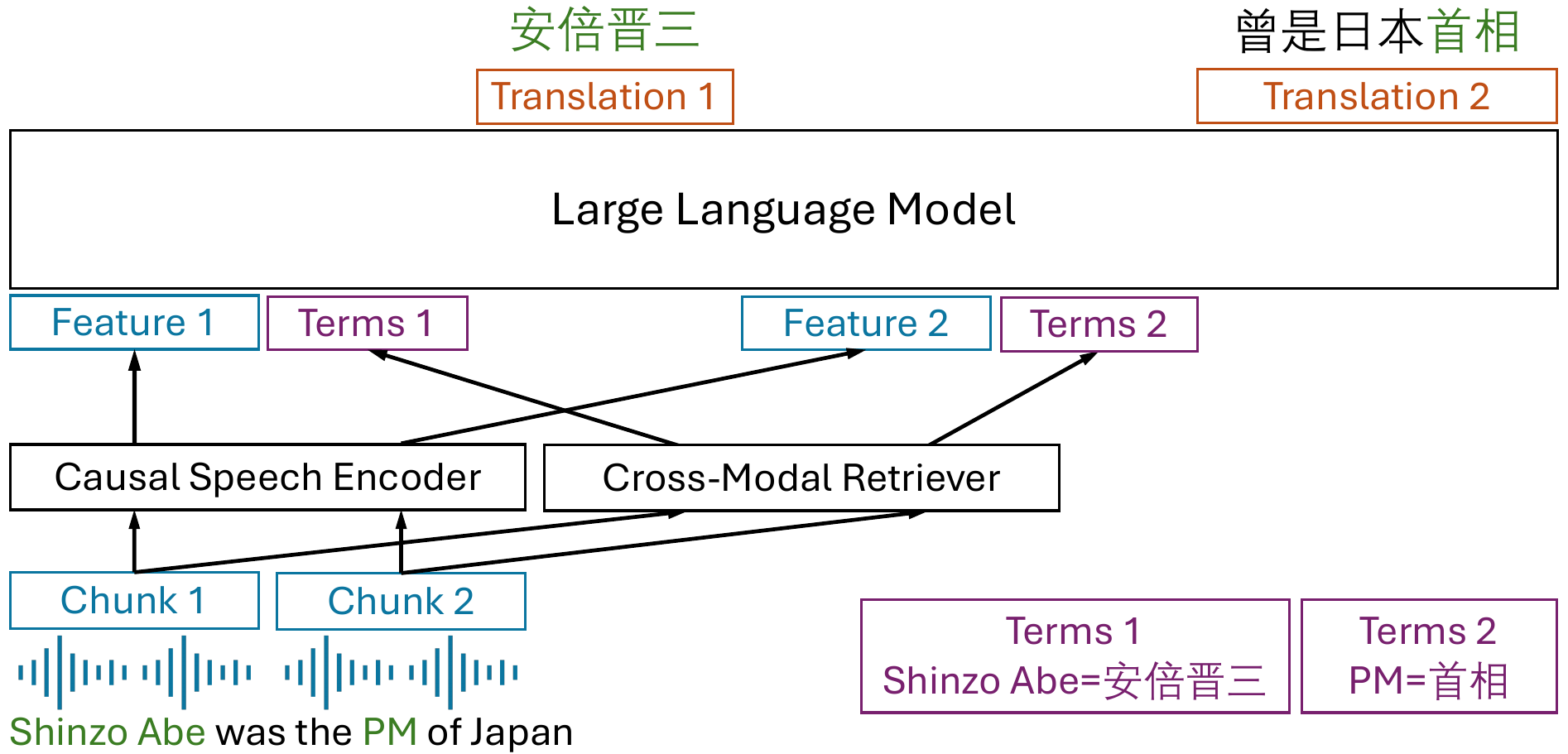}
    \caption{Overview of \method. \method interleaves speech chunks, retrieved terms, and translation outputs. For each incoming speech chunk, a causal speech encoder produces its feature representation, and a cross-modal retriever returns the corresponding terms. The retrieved terms are appended to the speech features and passed to the large language model, which decodes the next translation segment.}
    \label{fig:overview}
\end{figure*}

\section{Problem Formulation}
\label{sec:problem_formulation}

Let $\vs = (s_1, s_2, \dots, s_T) \in \mathbb{R}^T$ denote the source speech waveform, and let the glossary $\mathcal{G}=\{(e_j,e_j^\text{trans})\}_{k=1}^{|\mathcal{G}|}$, where $e_j$ is a term in the source language and $e_j^\text{trans}$ is its translation. At each step $i$, a fixed-duration speech chunk is received
\begin{align}
    \vs_i = \bigl(s_{i \cdot l + 1}, \dots, s_{(i+1) \cdot l}\bigr),
\end{align}
where $l$ is the chunk length.

Given the accumulated speech chunks $\vs_{1:i}$, the retriever $R_\phi$ selects a subset of relevant terminology entries
\begin{align}
    G_i = R_\phi(\vs_{1:i}, \mathcal{G}) \subseteq \mathcal{G} . \label{eqn:G_i}
\end{align}
Conditioned on the observed speech $\vs_{1:i}$, the previously generated translation tokens $\vy_{1:i-1}$, and the retrieved terminology context $G_{1:i}$, the model $\pi_\theta(\vs_{1:i}, \vy_{1:i-1}, G_{1:i})$ produces a partial translation $\vy_i$ at step $i$. The output $\vy_i = (y_1^i, \dots, y_{|\vy_i|}^i)$ may contain a variable number of tokens.

If $|\vy_i| = 0$, the model chooses to wait for additional speech input without emitting any translation. Otherwise, when $|\vy_i| > 0$, the model outputs a partial translation. All tokens in $\vy_i$ are assigned an identical delay of $i \cdot l$, corresponding to the elapsed time when the speech chunk $\vs_i$ becomes available.

Finally, translation quality and latency are evaluated with respect to the full source waveform $\vs$, the aggregated translation hypothesis $\vy$, the reference translation $\vy$, the source transcript $\vx$, and the delay associated with each generated token.

\section{Retrieval-Augmented Simultaneous Speech Translation}
\label{sec:method}

In this section, we first present an overview of RASST. We then describe the cross-modal retriever design and the training procedure that enables the Speech LLM to effectively leverage retrieved terms during simultaneous translation.



\subsection{\method Overview}
\label{sec:method_overview}

Figure \ref{fig:overview} presents an overview of \method. We adopt the interleaving SST formulation of InfiniSST~\citep{ouyang-etal-2025-infinisst}, which alternates between incoming speech chunks and incremental translation outputs, and \method augments this with a retrieval step:
\begin{align}
\vy_i &\sim \mathrm{SpeechLLM}(\vs_1, G_1, \vy_1, \ldots, \vs_i, G_i).
\end{align}
After each chunk $\vs_i$ arrives, the retriever returns candidate terms $G_i$ likely to appear in it, and we append these terms along with their target-language translations to the chunk's representation. Conditioned on the speech context and the retrieved terms, the LLM then generates the next partial translation $\vy_i$.

\subsection{Cross-Modal Retriever}
\label{sec:retriever}

\paragraph{Architecture}
We adopt a dual-encoder architecture. For text, we use the pre-trained BGE-M3 encoder~\citep{bge-m3} to map each source-language term $e_j$ to a feature vector $\vg_j\in\mathbb{R}^d$. For speech, we use the Audio Transformer of Qwen3-Omni~\citep{yang2025qwen3technicalreport} followed by a linear projection to the same dimension $d$. Since terms are typically short, the speech encoder reads only the current chunk $\vs_i$ together with a look-back window of duration $t_{lb}$, rather than the full history $\vs_{1:i}$ as in Equation \ref{eqn:G_i}. We denote this input as $\vs_i'$ and the resulting features as $\vf_i\in\mathbb{R}^{m\times d}$, with $m$ as the number of frames in $\vs_i'$.

\begin{figure}
    \centering
    \includegraphics[width=0.8\linewidth]{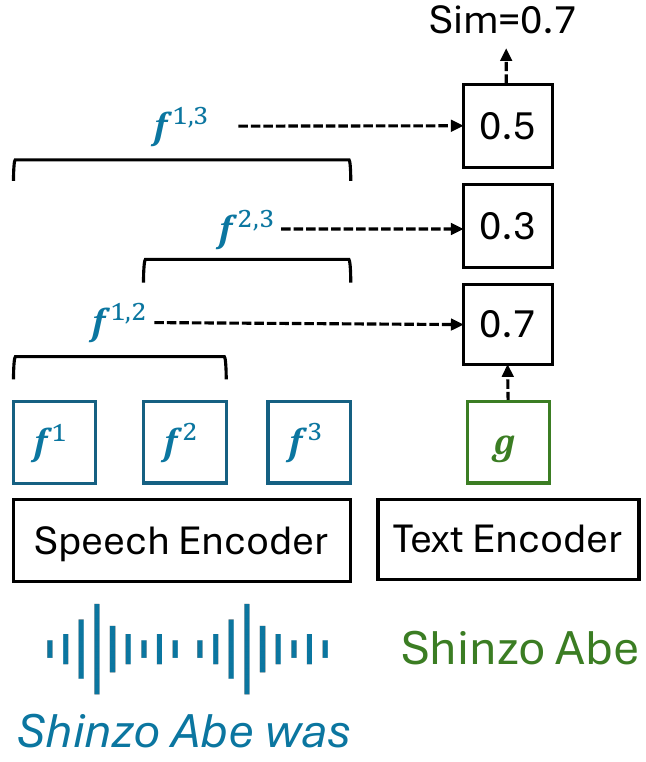}
    \caption{An example of multi-scale inference. The text term is encoded into a single feature vector $\vg$, while the speech context is encoded into frame features $\vf^1, \vf^2, \vf^3$. We enumerate all windows $(1,2), (2,3), (1,3)$ and compute the corresponding average-pooled features $\vf^{1,2}, \vf^{2,3}, \vf^{1,3}$. The likelihood that the term appears in the speech context is then the maximum cosine similarity between $\vg$ and any window feature.}
    \label{fig:multi_scale}
\end{figure}

\paragraph{Multi-Scale Inference}
Because terms vary in duration, representing the speech context with a single average-pooled vector loses information. Instead, we pool over windows at multiple scales. As shown in Figure \ref{fig:multi_scale}, given $\vf_i=(\vf_i^1,\vf_i^2,\ldots,\vf_i^m)$, for each window $1\leq u<v\leq m$ we compute
\begin{equation}
  \vf_i^{u,v} = \text{AvgPool}(\vf_i^u,\ldots,\vf_i^v)\in\mathbb{R}^d.
\end{equation}
The similarity between the speech $\vs_i'$ and a term $e_j$ is then the maximum cosine similarity over all windows:
\begin{equation}
  \text{Sim}(\vs_i',e_j)
  = \max_{1\leq u<v\leq m}
  \text{CosSim}(\vf_i^{u,v},\vg_j),
\end{equation}
and the retrieved terms are
\begin{align}
    G_i = \text{Top-}K_j\left(\text{Sim}(\vs_i',e_j)\right)\subseteq \mathcal{G}
\end{align}

In practice, we build a FAISS index~\citep{douze2024faiss} over the term vectors $\{\vg_j\}_{j=1}^{|\mathcal{G}|}$, then issue a Top-$K$ query for each window feature $\vf_i^{u,v}$, producing a candidate pool of size up to a multiple of $K$. We perform a second Top-$K$ selection over this pool by similarity score, keeping the highest score per term when duplicates occur, and finally discard any candidate whose similarity falls below a threshold $\tau$ to suppress noise.

\paragraph{Data Synthesis} \label{sec:retriever_data} Retriever training requires paired speech and terminology data $(\vs', e_\text{pos})$. Since terminology-rich speech corpora are scarce, we synthesize training data from two sources: existing ASR datasets and a terminology database.

For ASR data, we obtain word-level timestamps for each transcript using the Montreal Forced Aligner (MFA)\footnote{\url{https://github.com/MontrealCorpusTools/Montreal-Forced-Aligner}}, then extract noun phrases as candidate terms using spaCy\footnote{\url{https://spacy.io/} and  \texttt{en\_core\_web\_trf} model} and normalize their surface forms. We use noun phrases because most domain-specific terminology takes this form and they are abundant in standard transcripts. For each term, we carve out a speech context $\vs'$ of variable duration surrounding it, which gives the retriever flexibility across context lengths at inference time.

For each term in the terminology database, we prompt Gemini-2.5-Flash\footnote{\url{https://ai.google.dev/gemini-api/docs/models/gemini-2.5-flash}} to generate several sentences containing the term and synthesize the speech with CosyVoice 3~\citep{du2025cosyvoice}. The resulting speech is then processed by the same MFA pipeline as the ASR data.

\paragraph{Training} 
Given paired data $(\vs', e_\text{pos})$, we train the retriever with the InfoNCE loss~\citep{oord2019representationlearningcontrastivepredictive}. In addition to in-batch negatives, we mine hard negatives by using the retriever checkpoint, which is updated every 50 training steps, to retrieve the Top-$N$ most similar terms from the entire training set for each positive $e_\text{pos}$.
Let $E_\text{neg}$ be the negative set and $E_\text{all} = \{e_\text{pos}\} \cup E_\text{neg}$. The retriever loss $\mathcal{L}_{\mathrm{retriever}}$ is then \begin{equation} 
-\log \frac{ \exp(\text{Sim}(\vs_i',e_\text{pos})/\gamma) }{ \sum_{e\in E_\text{all}} \exp(\text{Sim}(\vs_i',e)/\gamma) }, 
\end{equation} where $\gamma$ is the temperature. 
Since the timestamp of $e_\text{pos}$ within $\vs'$ is known during training through MFA, we bypass the maximum over windows in $\text{Sim}(\cdot)$ and instead set $(u, v)$ to be the smallest window enclosing $e_\text{pos}$.

\subsection{Speech LLM Training}
\label{sec:rag_sst_generator}

To teach the Speech LLM to effectively use the retrieved terms $G_i
$ during simultaneous translation, we synthesize a corresponding training corpus. Given the speech chunks $\vs_{1:i}$, we follow InfiniSST's procedure to produce the translation segments $\vy_{1:i}$. The remaining piece beyond InfiniSST is the retrieved terms $G_{1:i}$. A naive approach is to reuse the noun-phrase terms from retriever training and always populate $G_i$ with the ground-truth term. This, however, would leave the model brittle to retriever mistakes at inference time. To reduce this training–inference mismatch, we build a glossary over all noun phrases in the training set and run the cross-modal retriever from Section~\ref{sec:retriever} on every speech chunk. Then we add the retrieved distractors in $G_i$ alongside the ground-truth term. The Speech LLM thus learns to translate in the presence of realistic retriever noise.

A training example is shown in Figure \ref{fig:train_prompt_visual}. We wrap the ground-truth term's translation in a \texttt{\textless term\textgreater\ldots\textless/term\textgreater} tag, encouraging the Speech LLM to copy ground-truth term translations directly from $G_i$. We fine-tune with a standard cross-entropy objective applied only to translation tokens $\vy$. Because translations naturally lag behind speech in the training data, the model also receives implicit supervision for \emph{timing}: it observes when a term appears in the speech context and when its translation appears later in the target sequence.
\section{Experimental Setups}
\label{sec:experiments}

\begin{figure*}[t]
\centering
\includegraphics[width=0.9\linewidth]{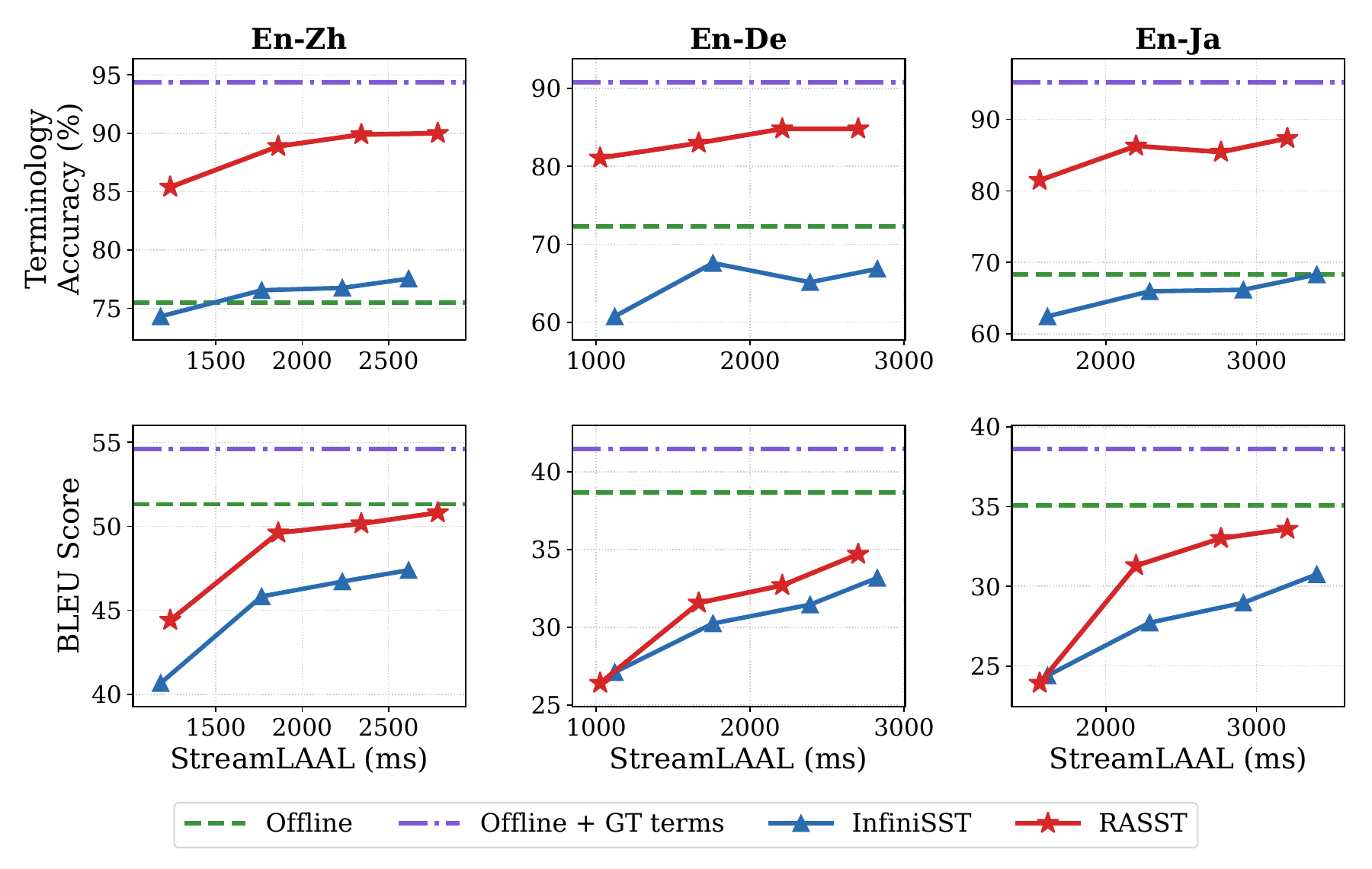}
\caption{Quality–latency trade-off of \method and baselines on the ACL 60/60 dev set. Quality is measured by both terminology accuracy and BLEU. Across all three language directions, \method improves terminology accuracy over InfiniSST by at least 10\% while also achieving higher translation quality.}
\label{fig:main_result_1}
\vspace{-0.5cm}
\end{figure*}

\subsection{Data}
\label{sec:task_datasets}

\paragraph{Retriever Training} We synthesize the retriever training data from GigaSpeech~\citep{chen21o_interspeech} and Wikidata\footnote{\url{https://www.wikidata.org/}}. GigaSpeech is a multi-domain English ASR corpus containing 10K hours of high-quality labeled audio. We process it following the procedure in Section~\ref{sec:retriever}, yielding 4.7M training pairs. The variable speech-context duration is sampled uniformly from ${2.88, 3.84, 4.80, 5.76}$ seconds, corresponding to the inference-time chunk sizes ${0.96, 1.92, 2.88, 3.84}$ plus a $t_\text{lb}=1.92$-second look-back. The look-back captures terms that cross chunk boundaries and we set it to $1.92$ seconds since most terms are shorter than this, as shown in Figure~\ref{fig:term_duration}.

For Wikidata, we extract candidate terms from the RDF truthy dump, keeping entities that have both an English \texttt{rdfs:label} and at least one \texttt{wdt:P31} triple. Full processing details are given in Appendix~\ref{sec:apdx_wikidata_proc}. For each retained term, we prompt Gemini-2.5-Flash to generate three natural utterances of 5--15 words containing the term (Figure \ref{fig:wikidata_utt_gen}). We then synthesize the utterances with CosyVoice 3\footnote{FunAudioLLM/Fun-CosyVoice3-0.5B-2512}, using speaker prompts randomly sampled from GigaSpeech. This results in 2.9M training pairs.

\paragraph{Speehc LLM Training}
We synthesize the Speech LLM training data using the same GigaSpeech ASR dataset. The interleaved En-Zh/De/Ja translation data is synthesized similar to InfiniSST with details in Appendix \ref{sec:speechllm_data_synth}. 
We subsample 12.5K instances for Speech LLM fine-tuning. We build a 140K glossary from GigaSpeech and Wikidata for distractor retrieval, and cap the size of $G_i$ at 20 terms.

\paragraph{Evaluation} We evaluate on two datasets. The first is the ACL 60/60 dev set~\citep{salesky-etal-2023-evaluating}, which contains five English ACL talks translated into 10 target languages. We evaluate on the En-Zh/De/Ja directions, feeding the unsegmented talks directly to the model. This dataset includes human-annotated terminology, which we use to construct the glossary for the model to retrieve from. After deduplication, the glossary contains 238 unique terms. The second is the ESO test set~\citep{eso_dataset}, which consists of five long-form pre-recorded English oncology training videos provided by the European School of Oncology, with translations into French, Spanish, German, and Slovene. We use GPT-5.4\footnote{\url{https://developers.openai.com/api/docs/models/gpt-5.4}} to supplement the dataset with Chinese and Japanese translations, and to extract a candidate terminology list. After manual verification, the resulting glossary contains 217 terms. To test robustness under larger glossaries, we expand both glossaries to 1K and 10K terms using Wikidata terms from domains similar to ACL or oncology and ensure that all added terms are unseen during training.

\subsection{Evaluation Metrics}
\label{sec:evaluation}

\paragraph{Latency}
Following IWSLT 2025 practice~\citep{agostinelli-etal-2025-findings}, we use StreamLAAL~\citep{papi-etal-2024-streamatt} as our latency metric. StreamLAAL segments the model hypothesis to align with reference sentences and reports the average latency over the resulting (hypothesis, reference) pairs.

\paragraph{Translation Quality}
We evaluate translation quality on the aligned (hypothesis, reference) sentence pairs using SacreBLEU~\citep{papineni-etal-2002-bleu,post-2018-call}.

\paragraph{Terminology Accuracy}
We introduce terminology accuracy to measure how accurately the model translates terms. For each terminology occurrence in the reference, we check whether its translation appears as an exact match in the aligned hypothesis. The metric is the percentage of such occurrences correctly matched, ranging from 0 to 100\%.

\subsection{Model Configuration}
\label{sec:retriever_training}

\paragraph{Architecture}
For the retriever, we use a Qwen3-Omni Audio Transformer\footnote{\url{https://huggingface.co/Atotti/Qwen3-Omni-AudioTransformer}} as the speech encoder, followed by a linear projection. For text encoding, we use the dense BGE-M3\footnote{\url{https://huggingface.co/BAAI/bge-m3}} encoder and extract the feature of \textsc{[CLS]} token. For the Speech LLM, we use Qwen3-Omni-30B-A3B-Instruct\footnote{\url{https://huggingface.co/Qwen/Qwen3-Omni-30B-A3B-Instruct}}.

\paragraph{Training}
We train both the cross-modal retriever and the Speech LLM with LoRA~\citep{hu2022lora}. For retriever training, the number of hard negative $N$ is 1024 and the global batch size is 8192. For Speech LLM training, we use sequence packing and a global batch size of 8192 tokens. Full hyperparameters and details are provided in Appendix~\ref{app:training_details}.

\paragraph{Inference}

We implement the inference agent in SimulEval~\citep{ma-etal-2020-simuleval}, varying the chunk size $l$ over ${0.96, 1.92, 2.88, 3.84}$ seconds. The Speech LLM is invoked once per chunk and generates up to 40 new tokens per call. We serve it with vLLM~\citep{kwon2023efficient} using tensor parallelism of size 2, decode with temperature $0.6$, top-$p{=}0.95$, and top-$k{=}20$, and strip the assistant's \texttt{<term>} tags from the hypothesis before scoring.

For retrieval, we build a FAISS inner-product index over $\ell_2$-normalized term embeddings. At each step, the retriever encodes the current speech chunk together with a $1.92$-second look-back, and queries the index using only those windows whose right boundary lies within the current chunk. We use Top-$K{=}10$ and discard candidates with similarity below $\tau{=}0.78$\footnote{We select $\tau$ on the dev set to allow at most a 1\% drop in recall.}.

\begin{figure*}[t]
\centering
\includegraphics[width=0.9\linewidth]{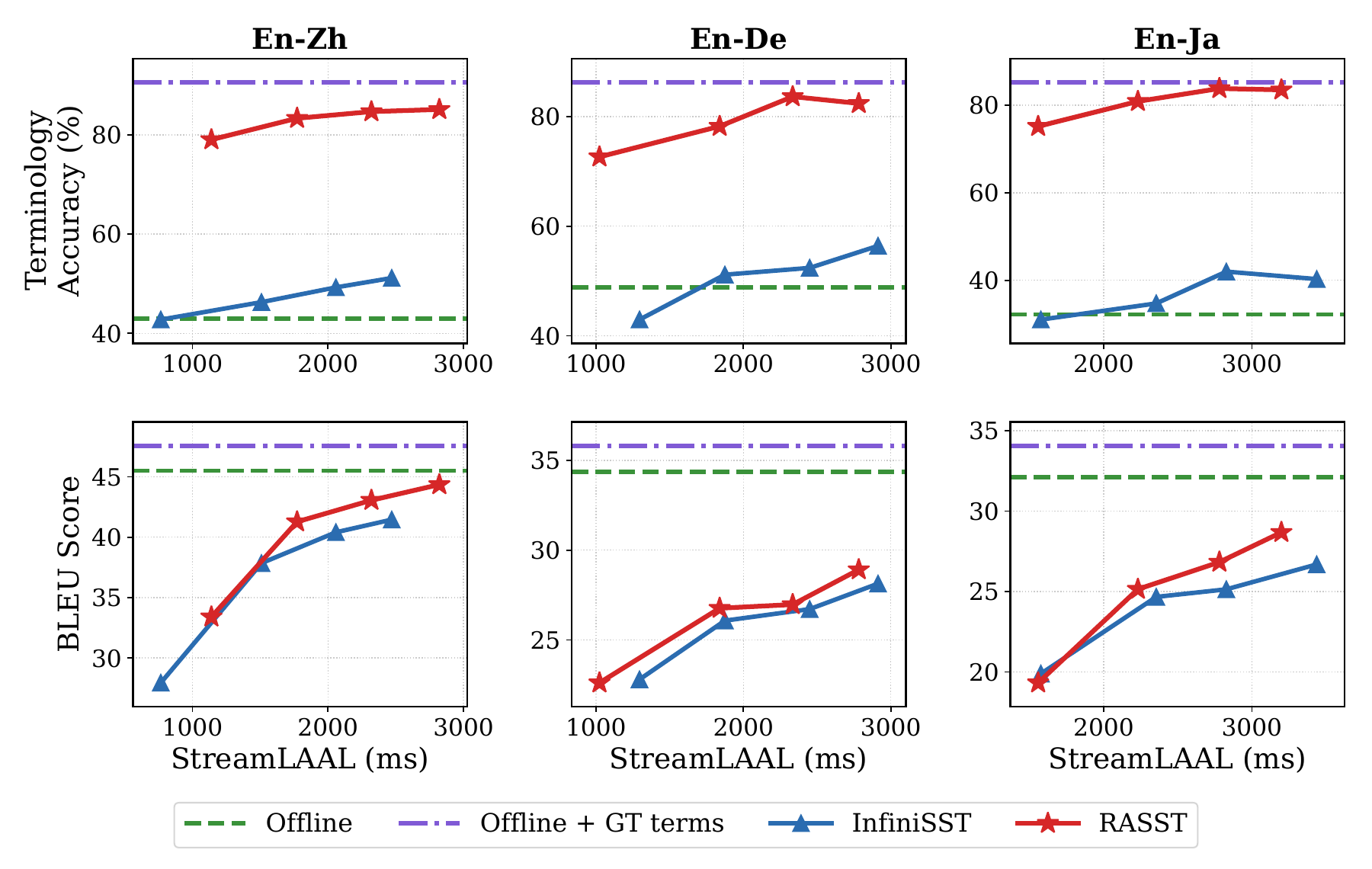}
\caption{Quality–latency trade-off of \method and baselines on the ESO test set. Quality is measured by both terminology accuracy and BLEU. Across all three language directions, \method improves terminology accuracy over InfiniSST by nearly 40\% while also achieving slightly higher translation quality.}
\label{fig:main_result_2}
\vspace{-0.5cm}
\end{figure*}

\begin{figure}[t]
\centering
\includegraphics[width=0.8\linewidth]{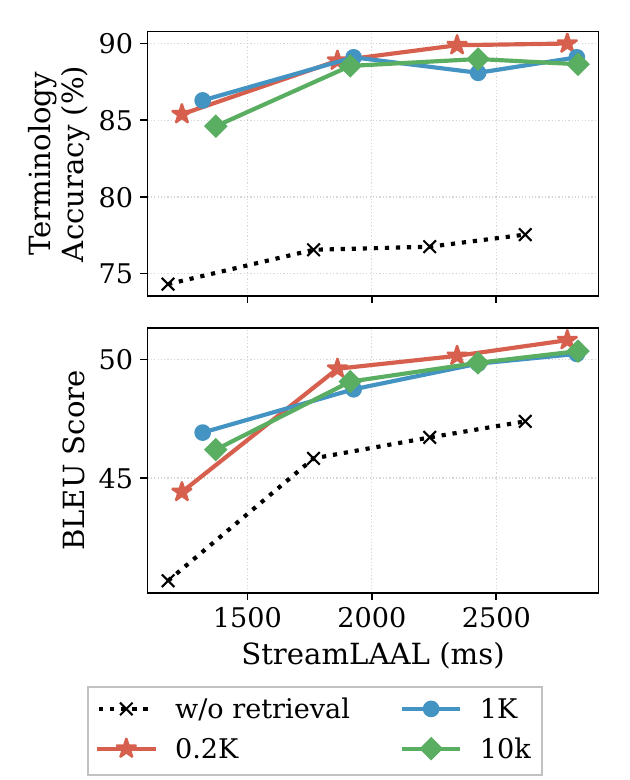}
\caption{Scaling the glossary from 0.2K to 1K and 10K terms. \method shows only a slight drop in terminology accuracy and translation quality, remaining well above the no-retrieval baseline.}
\label{fig:ablation_glossary_bank}
\vspace{-0.5cm}
\end{figure}


\subsection{Baselines}

\paragraph{Offline}
We use Qwen3-Omni-30B-A3B-Instruct to translate pre-segmented speech utterances offline, with the same decoding parameters as \method. This serves as an upper bound on translation quality.

\paragraph{Offline + GT Terms}
We extend the \textbf{Offline} baseline by adding the ground-truth terms appearing in each utterance to the prompt in a manner similar to \citet{gong24b_interspeech}, helping the model handle terminology correctly. This serves as an upper bound on terminology translation accuracy.

\paragraph{InfiniSST}
InfiniSST~\citep{ouyang-etal-2025-infinisst} is a strong SST baseline for simultaneous translation of unbounded speech and achieved top performance in the low-latency track at IWSLT 2025~\citep{ouyang-etal-2025-cmus,agostinelli-etal-2025-findings}. We train InfiniSST on the same data and with the same Speech LLM as \method, but remove all terminology input $G_i$. This setting can be viewed as \method without retrieval.

\section{Results and Analysis}


\subsection{Main Results}

\paragraph{\method substantially improves terminology accuracy}
As shown in the first row of Figures~\ref{fig:main_result_1} and~\ref{fig:main_result_2}, \method consistently improves terminology accuracy over the strong InfiniSST baseline, by at least 10\% on the ACL 60/60 dev set and nearly 40\% on the ESO test set. ESO sees a larger gain because its oncological terms are hard to recognize and translate, whereas many of the annotated terms in ACL 60/60 are everyday words like "words", "model", and "weights" which are easy to translate correctly. Notably, \method even surpasses the \textit{Offline} baseline, indicating that the problem cannot be solved simply by providing full speech context. A gap to \textit{Offline + GT Terms} remains because the retriever is imperfect, but it is typically under 5\%, underscoring \method's effectiveness.

\paragraph{\method improves overall translation quality}
As shown in the second row of Figures~\ref{fig:main_result_1} and~\ref{fig:main_result_2}, \method also improves overall translation quality over InfiniSST. The BLEU gains are smaller than the gains in terminology accuracy, which is expected since terms occur relatively infrequently in the speech.

\paragraph{\method is robust to large glossary size}
Figure~\ref{fig:ablation_glossary_bank} shows that \method remains competitive as the glossary scales to 10K terms. Going from 0.2K to 1K and then 10K terms gradually reduces terminology accuracy by less than 2\% and BLEU by less than 1 point, and \method still clearly outperforms the no-retrieval baseline.

\begin{table}[t]
\centering
\small
\begin{tabular}{lcccc}
\toprule
\textbf{Chunk Size (s)} & 0.96 & 1.92 & 2.88 & 3.84 \\
\midrule
Wall-Clock Time (s)    & 0.036 & 0.042 & 0.042 & 0.043 \\
\bottomrule
\end{tabular}
\caption{Retriever computation cost across speech chunk sizes. The retriever's overhead is negligible.}
\label{tab:rag_compute_rtf}
\vspace{-0.5cm}
\end{table}

\paragraph{\method is computationally efficient}
Table~\ref{tab:rag_compute_rtf} reports the additional computational cost introduced by the retriever. We run evaluation on two NVIDIA RTX A6000 GPUs, serving the Speech LLM with vLLM at tensor parallelism 2 and \texttt{gpu\_memory\_utilization}{=}0.72, while the retriever encoder fits in the remaining GPU memory. The retriever's wall-clock time per chunk is at most $3.75\%$ of the chunk size, and becomes smaller as the chunk grows, which makes it negligible in practice.

\subsection{Ablation Studies}
\label{sec:ablations}

\paragraph{Retriever hard negatives and threshold}
We examine the effect of training-time hard negatives and the inference-time threshold $\tau$. In addition to the default retriever trained with $N{=}1024$ hard negatives, we train two variants with $N{=}256$ and $N{=}0$. We evaluate their precision and recall on the GigaSpeech dev set against three glossaries: a 1K glossary built from GigaSpeech noun phrases, and 10K and 100K extensions with Wikidata terms unseen during training. For each (retriever variant, glossary) pair, we sweep $\tau$ from $0$ to $1$ to obtain a precision-recall curve. 

Figure~\ref{fig:ablation_hn_tau} shows the results. First, increasing $N$ yields clearly higher recall at the same precision, confirming the value of hard-negative mining. Second, recall degrades as the glossary grows. Based on these curves, we set $\tau{=}0.78$ at inference: it costs at most a 1\% drop in recall while roughly reducing the retrieval noise by half. Compared with $\tau=0$ in Figure \ref{fig:ablation_speechllm_tau}, $\tau=0.78$ achieves higher terminology accuracy. 

\paragraph{Multi-Scale inference}
We examine the contribution of multi-scale inference. As described in Section~\ref{sec:retriever}, \method's retriever queries with speech windows at multiple scales and keeps the terms with the highest similarities. We compare against two variants. The first reuses our retriever but queries with only the largest window at inference time, but this introduces a training–inference mismatch, since the retriever was trained with MFA-aligned windows around each term. The second variant removes this mismatch by training the retriever directly with the largest window, without MFA-based window localization. Results are shown in Figure \ref{fig:ablation_multiscale}. Multi-scale inference combined with MFA-localized training performs best: it loses less than 1\% in recall when scaling from 1K to 100K glossary, compared to drops of over 2\% and 5\% for the two variants.

\paragraph{Speech LLM Training Data}
We ablate the distractor-sampling strategy used to construct $G_i$ during Speech LLM training. Besides retriever mined distractors in \method, we train two variants: one with LLM-generated distractors, and one with no retrieved terms at all (equivalent to InfiniSST). At inference, all variants use the same retriever to populate $G_i$ from the speech chunk before the Speech LLM generates the translation. Results are shown in Figure \ref{fig:ablation_speechllm}. The model trained with LLM-generated distractors achieves slightly lower terminology accuracy and BLEU than \method. More surprisingly, \textit{InfiniSST + RAG}, which sees no retrieved terms during training but is given them at inference, matches \method's terminology accuracy but has much worse BLEU score. Inspecting its outputs, we find that it tends to copy retrieved terms verbatim without checking whether they actually appear in the source or inserting them at the right time.

More ablations on retriever training data, encoder choice and context length can be found in Appendix \ref{sec:apdx_ablation}.

\section{Conclusion}
We presented \method, a retrieval-augmented framework for simultaneous speech translation that improves domain-specific terminology translation. \method couples a lightweight cross-modal retriever with a Speech LLM and is trained with noise-robust supervision so that it remains accurate under realistic retrieval errors at inference time. On the ACL 60/60 dev set and the ESO test set, \method improves terminology accuracy by nearly 40\% and BLEU by up to 3 points over strong baselines.
\section*{Limitations}

This work has several limitations. 
First, we focus exclusively on English as the source language. Whether \method's cross-modal retriever and multi-scale inference transfer to other source languages remains untested. 
Second, our evaluation covers only two domains (academic talks and oncology training videos), leaving \method's effectiveness on other terminology-rich settings such as legal and financial domains uncharacterized. 
Third, we rely on a fixed similarity threshold $\tau$ to suppress retrieval noise.
Fourth, we do not explicitly model homophone confusion as in some prior contextual-biasing ASR work, since our focus is the integration of retrieval with an LLM-based simultaneous translation model, but it should be a natural extension. 
Finally, all experiments use Qwen3-Omni as the Speech LLM backbone, and we have not yet verified that our findings generalize to other speech LLMs.


\bibliography{custom,clean}

\appendix

\section{Wikidata Processing}
\label{sec:apdx_wikidata_proc}
We normalize each label by removing parenthetical qualifiers, lowercasing for deduplication, and retaining only short alphabetic or hyphenated terms with at most five words. We further require a non-empty short description from the glossary index that does not simply repeat the term, which filters out many ambiguous or non-English labels. To avoid a corpus dominated by very frequent instance types such as people, taxa, or scholarly articles, we assign each entity a primary \texttt{P31} type using its least frequent instance type and rank terms with inverse-type-frequency sampling. Terms appearing in the held-out evaluation glossaries are removed before training.

For each retained term, we prompt Gemini-2.5-Flash to generate three natural utterances of 5--15 words containing the term (Figure \ref{fig:wikidata_utt_gen}). We then synthesize the utterances with CosyVoice 3\footnote{\url{https://huggingface.co/FunAudioLLM/Fun-CosyVoice3-0.5B-2512}}, using speaker prompts randomly sampled from GigaSpeech.

\section{Speech LLM Data Synthesis}
\label{sec:speechllm_data_synth}

As in InfiniSST, we build robust segments of 28.8 seconds, filter out those with WER larger than 0.1 using parakeet-0.6b-tdt-v2\footnote{\url{https://huggingface.co/nvidia/parakeet-tdt-0.6b-v2}} and generate the target language translation with Qwen3-30B-A3B-Instruct-2507-FP8~\citep{yang2025qwen3technicalreport}\footnote{\url{https://huggingface.co/Qwen/Qwen3-30B-A3B-Instruct-2507-FP8}} and filter out those with MetricX score~\citep{juraska-etal-2024-metricx}\footnote{\url{https://huggingface.co/google/metricx-24-hybrid-xl-v2p6}} worse than -3. Finally we build the interleaved training data using SimAlign~\citep{jalili-sabet-etal-2020-simalign} with LaBSE~\citep{feng-etal-2022-language}. 

\begin{tcolorbox}[
    colback=gray!5,
    colframe=gray!50,
    title=Translation Prompt,
    fonttitle=\bfseries,
    boxrule=0.5pt
]
\ttfamily\small
You are given English document split into lines. Translate each line into Chinese. Do not include any other text.\\
\textless begin\textgreater\\
\{\}\\
\textless end\textgreater
\end{tcolorbox}

\section{Term Duration Distribution}
\label{app:term_duration}
\begin{figure}[t]
    \centering
    \includegraphics[width=0.9\linewidth]{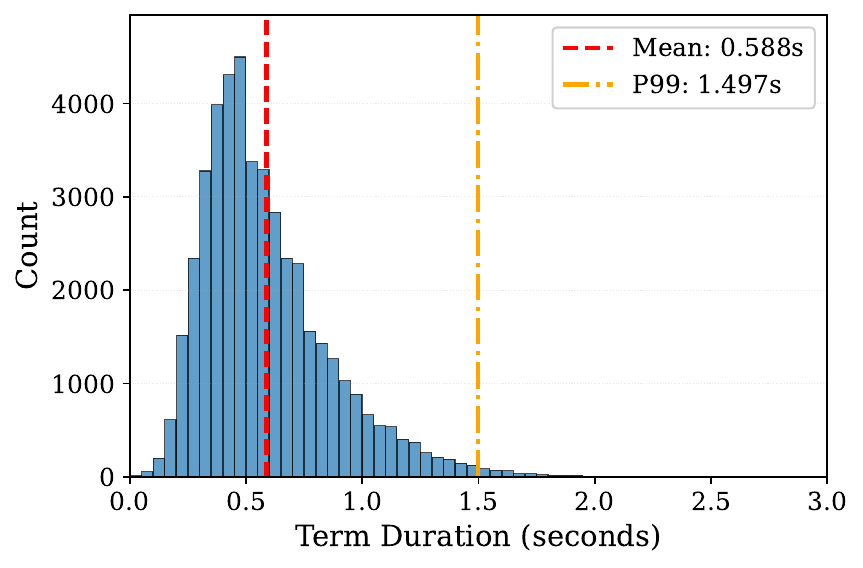}
    \caption{The distribution of term duration on GigaSpeech training set. }
    \label{fig:term_duration}
\end{figure}

We show the distribution of term duration on GigaSpeech training set in Figure \ref{fig:term_duration}.

\definecolor{trainframe}{RGB}{78, 86, 96}
\definecolor{trainsystembg}{RGB}{245, 248, 252}
\definecolor{trainuserbg}{RGB}{247, 250, 246}
\definecolor{trainassistantbg}{RGB}{255, 249, 239}
\providecommand{\appzh}[1]{\begin{CJK*}{UTF8}{gbsn}#1\end{CJK*}}
\newcommand{\trainbox}[2]{%
    \fcolorbox{trainframe}{#1}{%
        \begin{minipage}{\dimexpr\linewidth-2\fboxsep-2\fboxrule\relax}
        #2
        \end{minipage}%
    }%
}

\begin{figure}[t]
\centering
\includegraphics[width=0.8\linewidth]{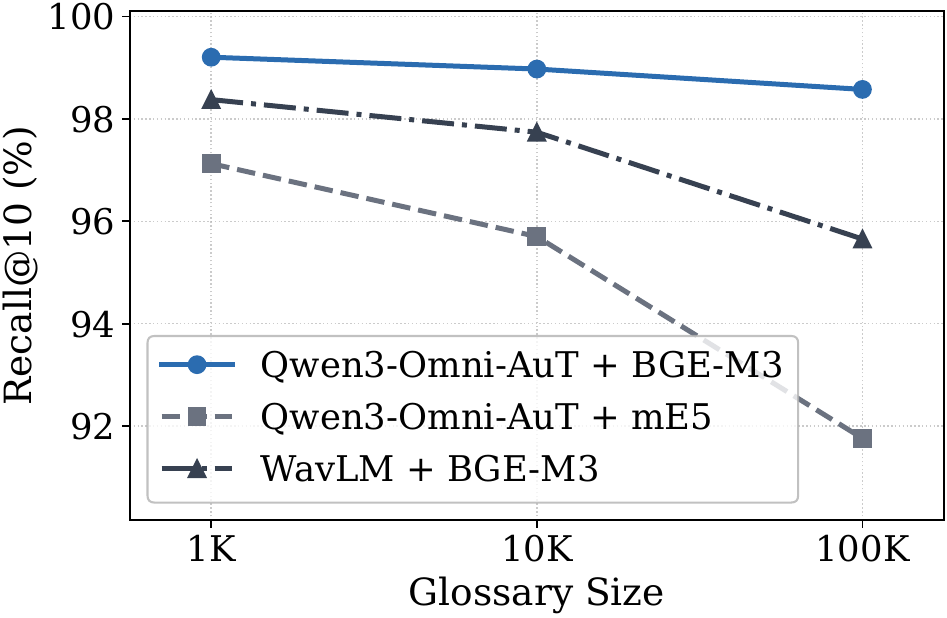}
\caption{Recall@10 across glossary sizes for different encoder combinations. Qwen3-Omni-AuT + BGE-M3 achieves the highest recall.}
\label{fig:ablation_retriever_encoder}
\end{figure}

\begin{figure}[t]
    \centering
    \includegraphics[width=\linewidth]{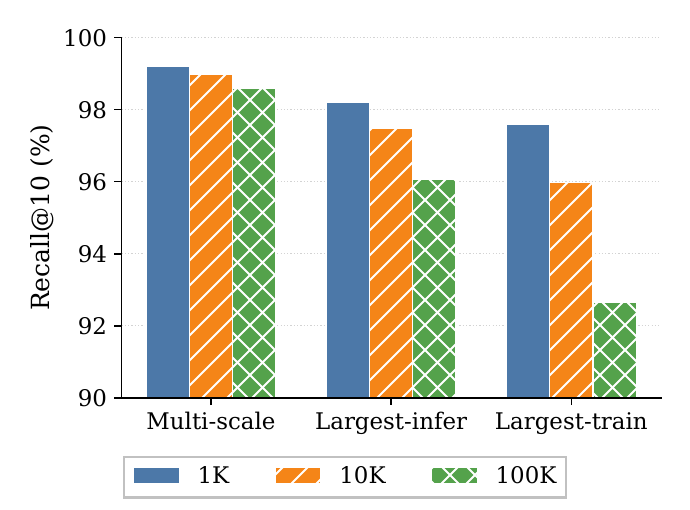}
    \caption{Recall@10 on the GigaSpeech dev set for multi-scale inference in \method and two variants. Largest-infer uses the same retriever as \method but queries with only the largest window at inference. Largest-train trains the retriever with the largest window instead of the MFA-localized window. Multi-scale inference combined with MFA-localized training performs best.}
    \label{fig:ablation_multiscale}
\end{figure}

\begin{figure}[t]
    \centering
    \includegraphics[width=0.8\linewidth]{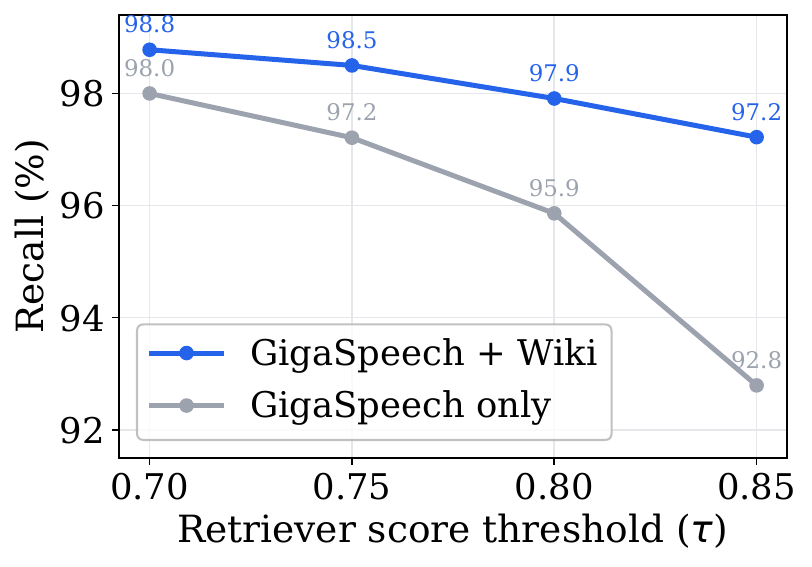}
    \caption{Recall@10 under different threshold $\tau$ on the GigaSpeech dev set for retrievers trained with and without Wikidata.}
    \label{fig:retriever_data}
\end{figure}

\begin{table}[t]
\centering
\setlength{\tabcolsep}{4pt}
\begin{tabular}{lccc}
\toprule
Context Length & 1K & 10K & 100K \\
\midrule
Fixed 1.92s & 98.11 & 97.47 & 95.18 \\
Fixed 3.84s & 98.65 & 98.31 & 97.58 \\
Fixed 5.76s & \textbf{99.64} & \textbf{99.58} & \textbf{99.33} \\
Variable 2.88--5.76s & 99.20 & 98.97 & 98.58 \\
\bottomrule
\end{tabular}
\caption{Recall@10 of retrievers trained with different context lengths, across glossary sizes. Longer speech contexts yield higher recall.}
\label{tab:ablation_context}
\end{table}

\section{Training Details}\label{app:training_details}

\paragraph{Retriever}
We train the retriever with LoRA applied to both the speech and text encoders.
For the audio encoder, LoRA is inserted into the query/key/value/output projections and MLP blocks with rank $r{=}32$ and scaling factor $\alpha{=}64$.
For the text encoder, LoRA targets the query/key/value projections with rank $r{=}16$ and $\alpha{=}32$.
We train with a multi-positive InfoNCE objective using a fixed temperature $\tau{=}0.03$.
We optimize with AdamW (learning rate $1\times10^{-4}$, weight decay 0.01), cosine annealing with 10\% warmup, and gradient clipping (max norm 1.0) under BF16 mixed precision.
Training is performed with Distributed Data Parallel (DDP) on eight RTX A6000 GPUs with a global batch size of 8192.

\paragraph{Speech LLM}
We fine-tune the Speech LLM using LoRA with rank $r{=}8$ and $\alpha{=}32$.
We enable sequence packing with a maximum sequence length of 2028 tokens.
We optimize with AdamW (learning rate $1\times10^{-4}$, weight decay 0.01) and gradient clipping (max norm 1.0), with a global batch size of 4.
Training is performed with DDP and expert parallelism (EP) of size 2 on two L40S GPUs.

\section{Additional Ablations}
\label{sec:apdx_ablation}

Additional ablations that cannot fit in the main text is here.

\begin{figure*}[t]
\centering
\includegraphics[width=0.9\linewidth]{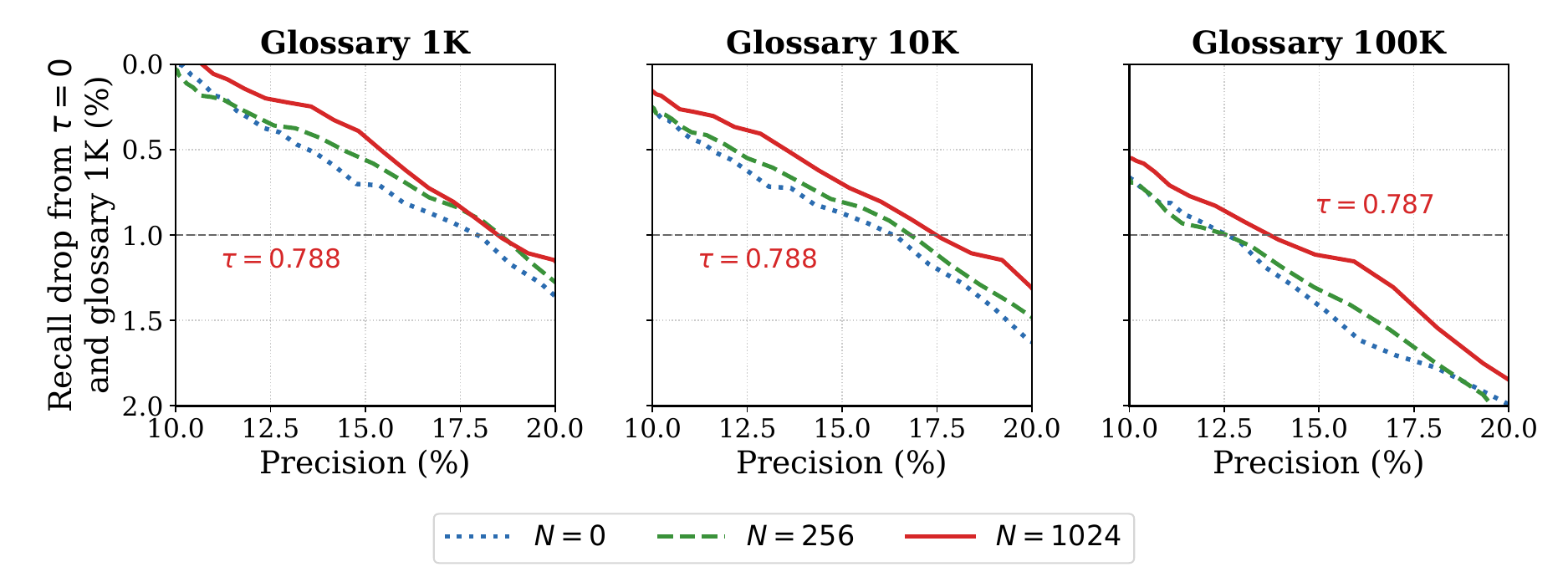}
\caption{Precision–recall curves across glossary sizes for retrievers trained with $N{=}0, 256, 1024$ hard negatives, obtained by sweeping $\tau \in [0,1]$. The dashed vertical line marks $\tau{=}0.78$, the operating point at which recall drops by 1\% from $\tau{=}0$ at the 1K glossary.}
\label{fig:ablation_hn_tau}
\end{figure*}

\begin{figure*}
    \centering
    \includegraphics[width=0.9\linewidth]{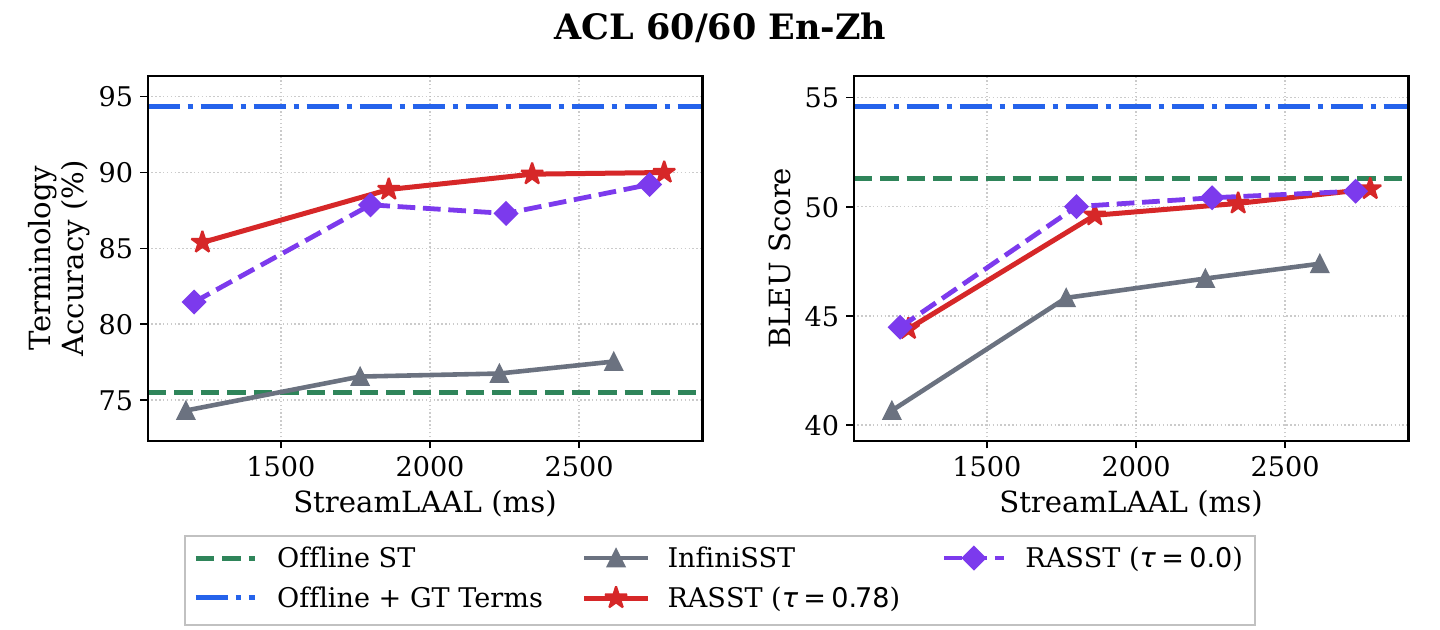}
    \caption{Quality–latency trade-off of \method with $\tau=0$ and $\tau=0.78$.} 
    \label{fig:ablation_speechllm_tau}
\end{figure*}

\begin{figure*}
    \centering
    \includegraphics[width=0.9\linewidth]{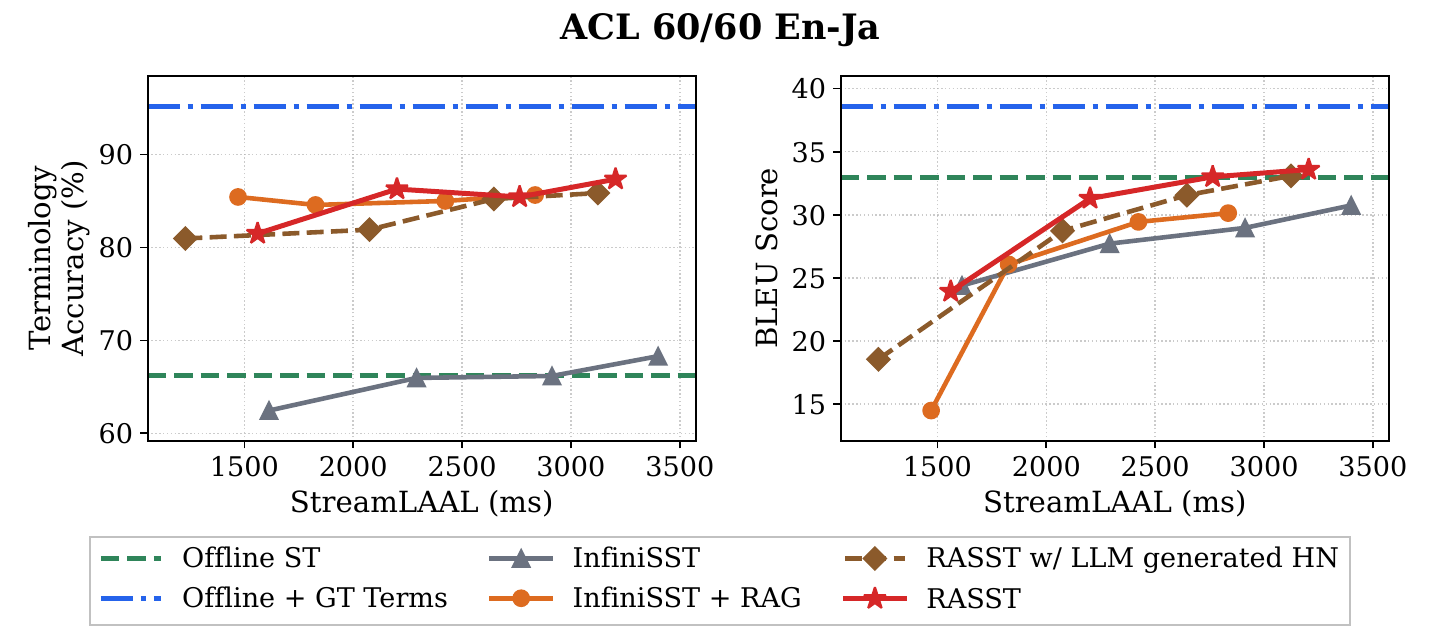}
    \caption{Quality–latency trade-off of \method against two training-data variants: a version trained with LLM-generated distractors, and \textit{InfiniSST + RAG}, which adds retrieval at inference to a model not trained with terms $G_i$. \method achieves the best overall trade-off.}
    \label{fig:ablation_speechllm}
\end{figure*}

\paragraph{Retriever training data} 
We examine the impact of training data by comparing our default retriever, trained on both GigaSpeech and Wikidata, against a GigaSpeech-only variant. As shown in Figure~\ref{fig:retriever_data}, adding Wikidata substantially improves recall, with the largest gains at high $\tau$, indicating that the retriever assigns higher similarity to true positives when trained with Wikidata.

\paragraph{Retriever encoders}
 We test different encoder combinations here. We additionally tried mE5 text encoder~\citep{wang2024multilingual}\footnote{\url{https://huggingface.co/intfloat/multilingual-e5-large}} and WavLM-large speech encoder~\citep{wavlm}\footnote{\url{https://huggingface.co/microsoft/wavlm-large}}. As shown in Figure~\ref{fig:ablation_retriever_encoder}, Qwen3-Omni-Aut + BGE-M3 achieves the best recall. 

\paragraph{Retriever context length.}
\method trains the retriever with variable-length speech contexts and here we compare against fixed-length alternatives. As shown in Table~\ref{tab:ablation_context}, longer fixed-length contexts generally yield higher recall across glossary sizes. The variable-length setup matches the recall of the longest fixed-length context (5.76s) while offering better flexibility at inference.

\begin{figure*}[t]
\centering
\begingroup
\setlength{\fboxsep}{5pt}
\setlength{\fboxrule}{0.35pt}
\renewcommand{\arraystretch}{1.08}
\scriptsize
\begin{tabularx}{\textwidth}{@{}>{\bfseries}p{0.12\textwidth}>{\raggedright\arraybackslash}X@{}}
System & \trainbox{trainsystembg}{You are a professional simultaneous interpreter. You will be given chunks of English audio and you need to translate the audio into Chinese text. Use the \texttt{term\_map} as a reference for terminology if provided.} \\
\addlinespace[0.45ex]
User 1 & \trainbox{trainuserbg}{\texttt{<audio>} \par \textbf{term\_map:} \texttt{things} $\rightarrow$ \appzh{事}; \texttt{everything} $\rightarrow$ \appzh{所有努力}; \texttt{right} $\rightarrow$ \appzh{权利}; \texttt{thing} $\rightarrow$ \appzh{东西}; \texttt{that thing} $\rightarrow$ \appzh{那东西}.} \\
Assistant 1 & \trainbox{trainassistantbg}{\appzh{她会做些我本意不想让她做的}\texttt{<term>}\appzh{事}\texttt{</term>}\appzh{。这样一来，之前}\texttt{<term>}\appzh{所有努力}\texttt{</term>}\appzh{都白费了，我不得不重新写一遍。}} \\
\addlinespace[0.45ex]
User 2 & \trainbox{trainuserbg}{\texttt{<audio>} \par \textbf{term\_map:} \texttt{the Porch Gable} $\rightarrow$ \appzh{门廊小屋}; \texttt{story} $\rightarrow$ \appzh{传说}; \texttt{Anne} $\rightarrow$ \appzh{安妮}; \texttt{seclusion} $\rightarrow$ \appzh{静谧}; \texttt{Diana} $\rightarrow$ \appzh{戴安娜}; \texttt{gabelle} $\rightarrow$ \appzh{法国盐税}; \texttt{that story} $\rightarrow$ \appzh{这个故事}; \texttt{porch} $\rightarrow$ \appzh{门廊}.} \\
Assistant 2 & \trainbox{trainassistantbg}{\appzh{然而，最终}\texttt{<term>}\appzh{传说}\texttt{</term>}\appzh{终于完成了，}\texttt{<term>}\appzh{安妮}\texttt{</term>}\appzh{在}\texttt{<term>}\appzh{门廊小屋}\texttt{</term>}\appzh{的}\texttt{<term>}\appzh{静谧}\texttt{</term>}\appzh{中把它读给了}\texttt{<term>}\appzh{戴安娜}\texttt{</term>}\appzh{听。她}} \\
\addlinespace[0.45ex]
User 3 & \trainbox{trainuserbg}{\texttt{<audio>} \par \textbf{term\_map:} \texttt{Diana} $\rightarrow$ \appzh{戴安娜}; \texttt{watchful eye} $\rightarrow$ \appzh{密切留意}; \texttt{Robert Ray} $\rightarrow$ \appzh{罗伯特·雷}; \texttt{her pathetic scene} $\rightarrow$ \appzh{那个悲情场景}; \texttt{scene} $\rightarrow$ \appzh{现场}; \texttt{that scene} $\rightarrow$ \appzh{那场戏}.} \\
Assistant 3 & \trainbox{trainassistantbg}{\appzh{成功地呈现了那个悲情场景，同时又没有牺牲}\texttt{<term>}\appzh{罗伯特·雷}\texttt{</term>}\appzh{，她一边读一边}\texttt{<term>}\appzh{密切留意}\texttt{</term>}\appzh{着}\texttt{<term>}\appzh{戴安娜}\texttt{</term>}\appzh{的反应。戴安娜}} \\
\end{tabularx}
\endgroup
\caption{An example of En-Zh Speech LLM training data.}
\label{fig:train_prompt_visual}
\end{figure*}

\begin{figure*}[t] 
\begin{tcolorbox}[
    enhanced,
    colback=gray!3,
    colframe=gray!55!black,
    title={Prompt template for Wikidata term utterance generation},
    fonttitle=\bfseries,
    boxrule=0.4pt,
    arc=1mm,
    left=1.5mm,
    right=1.5mm,
    top=1mm,
    bottom=1mm,
    width=\textwidth 
]
{\ttfamily\footnotesize
\textbf{System instruction:}\par
You are a dataset builder for speech recognition research.\par
Your task: given a list of terms with their short descriptions, generate short English utterances (5--15 words each) that a speaker might say in a lecture, tutorial, or technical discussion. Use the description as context to produce accurate, meaningful sentences, but do NOT just copy the description.\par
Each utterance MUST contain the given term EXACTLY as written (same spelling, but case may differ). Vary sentence structure and term position (beginning, middle, end). Do NOT repeat the same pattern across terms.\par
Output valid JSON only -- no markdown fences, no explanation.\par
\vspace{0.5em}
\textbf{User prompt:}\par
Generate \{variants\_per\_term\} short utterances (5--15 words) for EACH of the following terms. Use the description as context to make the utterance accurate and meaningful. Each utterance must contain the term verbatim.\par
\vspace{0.5em}
Terms with descriptions:\par
[\par
\hspace*{1em}\{"term": "Seismology", "description": "scientific study of earthquakes"\},\par
\hspace*{1em}\{"term": "dataflow graph", "description": "graph representation of data dependencies"\}\par
]\par
\vspace{0.5em}
Return a JSON object mapping each term to a list of \{variants\_per\_term\} utterance strings. Example format:\par
\{\par
\hspace*{1em}"Seismology": [\par
\hspace*{2em}"Seismology helps us understand earthquake patterns.",\par
\hspace*{2em}"The field of seismology has advanced rapidly."\par
\hspace*{1em}]\par
\}\par
\vspace{0.5em}
IMPORTANT: output raw JSON only, no markdown code fences.\par
}
\end{tcolorbox}
\caption{Prompt template for Wikidata term utterance generation.}
\label{fig:wikidata_utt_gen}
\end{figure*}

\end{document}